\documentclass[runningheads, dvipsnames]{llncs}
\usepackage{cite}
\usepackage{svg}
\usepackage{color}
\usepackage{mathtools}
\usepackage{graphicx}
\usepackage{tabularx, booktabs}
\usepackage{caption, subcaption}
\usepackage{varwidth}
\usepackage{enumerate}
\usepackage[inline]{enumitem}
\usepackage{multirow}
\usepackage[T1]{fontenc}
\usepackage{amsmath, amssymb}
\usepackage{xcolor}

\makeatletter
\def\set@curr@file#1{\def\@curr@file{#1}}
\makeatother 

\begin{document}
\title{Identifying Suitable Tasks for Inductive Transfer Through the Analysis of Feature Attributions}
%
\titlerunning{Leveraging Feature Attributions for Transfer Learning}
%
\author{Alexander Pugantsov\orcidID{0000-0001-6630-0258} \\
Richard McCreadie\orcidID{0000-0002-2751-2087}}
\authorrunning{A. J. Hepburn and R. McCreadie}
%
\institute{University of Glasgow, University Avenue, Glasgow, G12 8QQ, Scotland\\\email{a.pugantsov.1@research.gla.ac.uk\\ richard.mccreadie@glasgow.ac.uk}}
%
\maketitle              

\newcommand{\todo}[1]{\textcolor{red}{TODO:[#1]}}
\newcommand{\ric}[1]{\textcolor{black}{#1}}
\newcommand{\revisions}[1]{\textcolor{black}{#1}}

\begin{abstract}
Transfer learning approaches have shown to significantly improve performance on downstream tasks. However, it is common for prior works to only report where transfer learning was beneficial, ignoring the significant trial-and-error required to find effective settings for transfer. Indeed, not all task combinations lead to performance benefits, and brute-force searching rapidly becomes computationally infeasible. Hence the question arises, \emph{can we predict whether transfer between two tasks will be beneficial without actually performing the experiment?} In this paper, we leverage explainability techniques to effectively predict whether task pairs will be complementary, through comparison of neural network activation between single-task models. In this way, we can avoid grid-searches over all task and hyperparameter combinations, dramatically reducing the time needed to find effective task pairs. Our results show that, through this approach, it is possible to reduce training time by up to 83.5\% at a cost of only 0.034 reduction in positive-class F1 on the TREC-IS 2020-A dataset.

\end{abstract}

\section{Introduction}
Transfer learning is a method of optimisation where models trained on one task are repurposed for another downstream task. The intuition behind this approach is clear; as human beings, we often apply knowledge learned from previous experience when learning a new, related skill. Hence, transfer learning aims to mimic this biological behaviour by exploiting the relatedness between tasks.

However, there remains an ever-present question that researchers have long strived to answer, \textit{Why is pretraining useful for my task?} More specifically, \textit{What information encoded in a pretrained model is transferrable for my task?} If, hypothetically, we are capable of approximating, prior to training, which auxiliary tasks will be useful in practice, we are then able to avoid the often laborious process of trial-and-error over all task and parameter combinations. Hence, we propose a solution which leverages recent research in explainability to identify the properties that characterise particular tasks and by extension, the properties which make these tasks related.

Through the evaluation of 803 models, we calculate the per-document term activity for each task and use these to predict the performance outputs of each combined task pair. We show that there exists correlation between strongly-attributed shared terms between pairs of single tasks and their combined performance output, and that, by ranking each task pair by their performance, we can reduce the time it takes to find the best-performing model by up to 83.5\% (with a cost of only 0.034 reduction in positive-class F1).

\section{Improving Performance Through Inductive Transfer}
\noindent The concept of \textit{inductive transfer}, introduced by Pan and Yang~\cite{Pan2010ASO}, can be considered a method of transfer learning wherein the source and target tasks are different, the goal of which is to leverage domain information in the source task---encoded in the training signals as an inductive bias---to be transferred to a downstream, target task. 

However, the necessary conditions for what constitutes a suitable auxiliary task for use in pretraining is unclear. Mou et al.~\cite{mouHowTransferableAre2016} note that the difficulty in transferability in this domain lies in the discreteness of word tokens and their embeddings. Similar to this work, Bingel and S{\o}gaard~\cite{bingelIdentifyingBeneficialTask2017} identified beneficial task relations for multi-task learning and found that performance gains were predictable from the dataset characteristics. While ground has been covered in understanding and quantifying the relationship between pairs of tasks, what constitutes task relatedness remains an open question. To this end, we first demonstrate the efficacy of transfer learning as a method of improving classifier performance. We utilise the dataset provided by the TREC Incident Streams Track (TREC-IS) which features a number of multi-label classification tasks wherein each label is representative of some information need (known as \textit{information types}) to end users of automated crisis and disaster systems. More importantly, these labels exhibit some level of conceptual relatedness, and as such, is an appropriate framework for this investigation. The track features 25 labels which manual assessors may ascribe to each document, however, to limit the number of models trained, we use the track's \textbf{Task 2} formulation, which restricts the number of information types to 12\footnote{More information on metrics and tasks can be found at \url{http://trecis.org}}.


We experiment with transfer learning across these information types, that is to say, we train a particular classifier on one, \textit{source} task and then use the resulting model as a pretrained baseline for tuning another downstream \textit{target} task, using a pretrained BERT transformer model as defined by Devlin et al.~\cite{Devlin2019BERTPO} as the base model for our experiments.

\begin{table}[t]
\renewcommand{\arraystretch}{1.2}%
\resizebox{120mm}{!}{
    \begin{tabular}{|l|l|l|c|c|c|c|c|c|c|c|}
\hline
\multicolumn{2}{|c|}{ } & \multicolumn{4}{c|}{Inductive Transfer (Source)} & \multicolumn{3}{c|}{Target Parameters} &\multicolumn{2}{c|}{Evaluation Scores} \\
\cline{1-11}
Target & Model & Transfer-From & LR & \#E & B\# & LR & \#E & B\#  & Positive F1 & Accuracy\\
\hline
\hline
\multirow{2}{*}{New Sub Event} & BERT$\rightarrow$Target & None & - & - & - & 2e-05 & 4 & 16 & 0.0258 & 0.9604 \\
\cline{2-11}
& BERT$\rightarrow$Source$\rightarrow$Target & Other (Best) & 1e-05 & 2 & 32 & 2e-05 & 2 & 32 & 0.0578 & 0.9432 \\
\hline
\hline
\multirow{2}{*}{First Party Observation} & BERT$\rightarrow$Target & None & - & - & - & 2e-05 & 4 & 32 & 0.0259 & 0.9646 \\
\cline{2-11}
& BERT$\rightarrow$Source$\rightarrow$Target & Move People (Best) & 1e-05 & 2 & 32 & 1e-05 & 1 & 32 & 0.1142 & 0.9538 \\
\hline
\hline
\multirow{2}{*}{Service Available} & BERT$\rightarrow$Target & None & - & - & - & 3e-05 & 3 & 16 & 0.0944 & 0.9821 \\
\cline{2-11}
& BERT$\rightarrow$Source$\rightarrow$Target & Other (Best) & 1e-05 & 1 & 32 & 1e-05 & 1 & 32 & 0.1095 & 0.9783 \\
\hline
\hline
\multirow{2}{*}{Move People} & BERT$\rightarrow$Target & None & - & - & - & 2e-05 & 3 & 32 & 0.1964 & 0.9835 \\
\cline{2-11}
& BERT$\rightarrow$Source$\rightarrow$Target & Other (Best) & 1e-05 & 1 & 32 & 1e-05 & 2 & 32 & 0.2423 & 0.9853 \\
\hline
\hline
\multirow{2}{*}{Emerging Threats} & BERT$\rightarrow$Target & None & - & - & - & 3e-05 & 2 & 32 & 0.2329 & 0.8323 \\
\cline{2-11}
& BERT$\rightarrow$Source$\rightarrow$Target & Location (Best) & 1e-05 & 2 & 32 & 1e-05 & 1 & 32 & 0.2612 & 0.8135 \\
\hline
\hline
\multirow{2}{*}{Multimedia Share} & BERT$\rightarrow$Target & None & - & - & - & 2e-05 & 3 & 32 & 0.4356 & 0.6760 \\
\cline{2-11}
& BERT$\rightarrow$Source$\rightarrow$Target & Other (Best) & 1e-05 & 2 & 32 & 2e-05 & 1 & 32 & 0.4709 & 0.6422 \\
\hline
\hline
\multirow{2}{*}{Location} & BERT$\rightarrow$Target & None & - & - & - & 3e-05 & 2 & 16 & 0.5904 & 0.6939 \\
\cline{2-11}
& BERT$\rightarrow$Source$\rightarrow$Target & Multimedia Share (Best) & 1e-05 & 1 & 32 & 1e-05 & 1 & 32 & 0.6178 & 0.7196 \\
\hline
\hline
\multirow{2}{*}{Other} & BERT$\rightarrow$Target & None & - & - & - & 5e-05 & 4 & 16 & 0.6831 & 0.5638 \\
\cline{2-11}
& BERT$\rightarrow$Source$\rightarrow$Target & Multimedia Share (Best) & 2e-05 & 1 & 32 & 1e-05 & 2 & 32 & 0.6853 & 0.7187 \\
\hline
\multicolumn{11}{c}{}\\
\hline
\multirow{2}{*}{AVERAGE} & BERT$\rightarrow$Target & None & \multicolumn{3}{c|}{-} & \multicolumn{3}{c||}{\emph{Varies}} & 0.2856 & 0.8321 \\
\cline{2-11}
& BERT$\rightarrow$Source$\rightarrow$Target & \emph{Varies} & \multicolumn{3}{c|}{\emph{Varies}} & \multicolumn{3}{c||}{\emph{Varies}} & 0.3199 & 0.8443 \\
\hline
    \end{tabular}}
    \caption{\small{Information type categorisation performance with and without inductive transfer from a source task. Metrics are micro-averaged across events and range from 0 to 1, higher is better.}}
    \label{table:transfer-results}
    \vspace{-9mm}
\end{table}

Table~\ref{table:transfer-results} shows the single- and multi-task model results from previous experiments, containing each task's baseline performance (omitting 4 tasks which showed no performance change) and their respective best-performing auxiliary task when used as a prior. With the exception of those omitted tasks, we observed performance increases across the board\ric{, as can be seen from comparing the BERT$\rightarrow$Target and BERT$\rightarrow$Source$\rightarrow$Target rows for each task in the above table. However, obtaining these improvements was not a trivial process.} We found that performance increases were highly dependent on the target information type \ric{and that the effectiveness of transfer was highly sensitive to changes in model hyperparameters. Moreover, there were no easily discernible patterns that we could use as heuristics to speed up the process of finding the best model, leading to an exhaustive grid-search over all task and parameter combinations, calling into question the practicality of such an approach in production. Hence, if we are to realise these performance gains, a cheaper approach to finding effective pairs of tasks is needed.}

\section{Optimising Transfer Learning with Explainability}
\noindent As the complexity of deep neural models grows exponentially, there is an increasing need for methods to enable a deeper understanding of the latent patterns of a neural model, such as when trying to understand cases where that model has failed. In order to understand this behaviour, we must explore methods of explaining the inner working of language models.

\textit{Explainability} is a field focused on model understanding and the predictive transparency of machine learning-based systems. A number of explainability techniques take the form of gradient-based approaches~\cite{DBLP:journals/corr/RibeiroSG16, mundhek2019saliency}. One such gradient-based approach, known as \textit{attribution}-based explanations, allow us to assess what the dominant features were that contributed to a particular prediction. Various algorithms can assign an importance score to each given input feature and effectively summarise and visualise these scores in a human-readable manner. Attribution-based explainability has become especially popular in the literature~\cite{liu2020ExplainableRecommender, ismail2020explainablebenchmark, zhang2021explainpredict, Wu2020explainingquestions, zeiler2013convexpl}, however, research into explainability for transfer learning is sparse.

In this work, we investigate: 1) whether there exists correlation between the shared, \textit{important} linguistic properties of a pair of tasks and their combined performance output; 2) whether we can compute this relationship prior to training these combined models; and 3) whether we can, as a result, reduce the time taken to produce high-performance models. As such, we divide the remainder of this paper into the following research questions:

\begin{enumerate}[
    leftmargin=1.5cm,
    label={\textbf{RQ\arabic*}.}]
    \item Does there exist some degree of correlation between the shared, \textit{active} terms between pairs of tasks and their combined performance output?
    \item Can we leverage this knowledge, prior to training, to reduce the overall runtime required to produce \ric{effective} models?
\end{enumerate}

To this end, we compute the \textit{conductance} of latent features in the context of each document. Introduced by Dhamdhere et al.~\cite{dhamdhere2017ig, dhamdhere2018conductance} the conductance of a hidden unit can be described as the flow of attributions via said unit. By computing the conductance, we are able to quantify the bearing each individual input feature has on a particular prediction (with respect to a given input sequence).

\begin{figure}[t]
\centering
\includegraphics[scale=0.45]{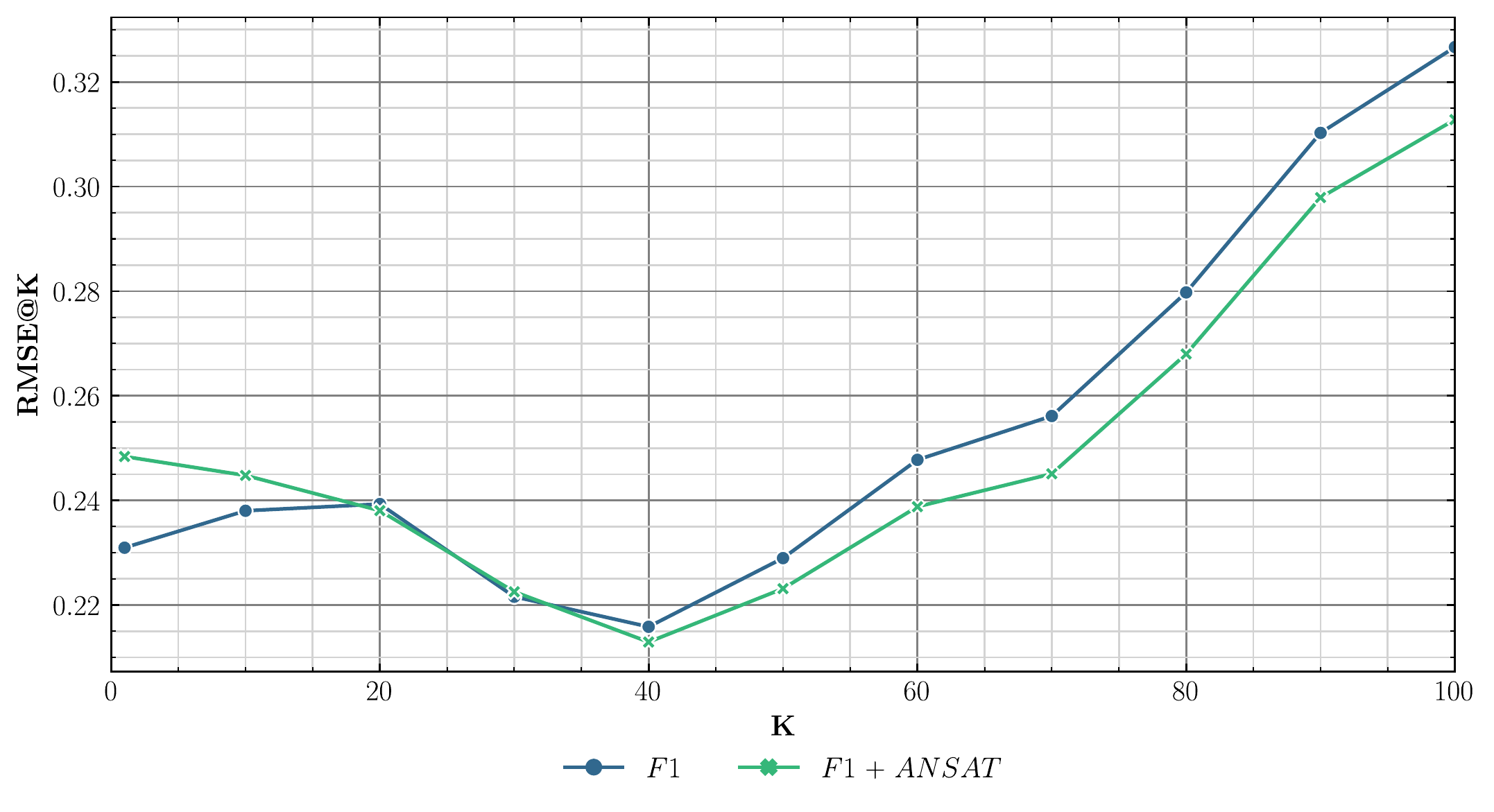}
\vspace{-2mm}
\caption{\small{RMSE@k results from our XGBoost regression model. RMSE metrics range from 0 to $\infty$, lower is better.}}\label{fig:rmse_at_k}
\vspace{-5mm}
\end{figure}

For each \ric{BERT$\rightarrow$Target} model and each document in our test set, we calculate the effect any individual feature \ric{(term)} had on the prediction output of its document using conductance. The conductance $c$ of each term $x_{i}$ within a document is scored $\{c_{x_{i}} \in \mathbb{R} : -1 \leq c_{x_{i}} \leq 1\}$ wherein $c_{x_{i}} \in [-1, 0)$ represents conductance scores that attribute towards the negative class and $c_{x_{i}} \in (0, 1]$ attribute towards our target class. We eliminate negatively attributed terms in order to capture the most \textit{active} terms that represent our target class. We determine activity by testing against a range of thresholds for term activity ($TAT$), beginning from the mean of positively-attributed conductance scores, 0.05, and increasing to a reasonable upper bound at 0.05 increments. As such, we decided to test the set of conductance thresholds: \revisions{$[0.05, 0.7]\cap0.05\mathbb{Z}$.} We then averaged the total number of active terms across documents and only consider those terms which are above said thresholds. Our calculations result in the following formulation, Average Number of Shared Active Terms ($ANSAT$), which provides a quantified comparison metric between each pair of models:

\begin{figure}[t]
\centering
\includegraphics[scale=0.45]{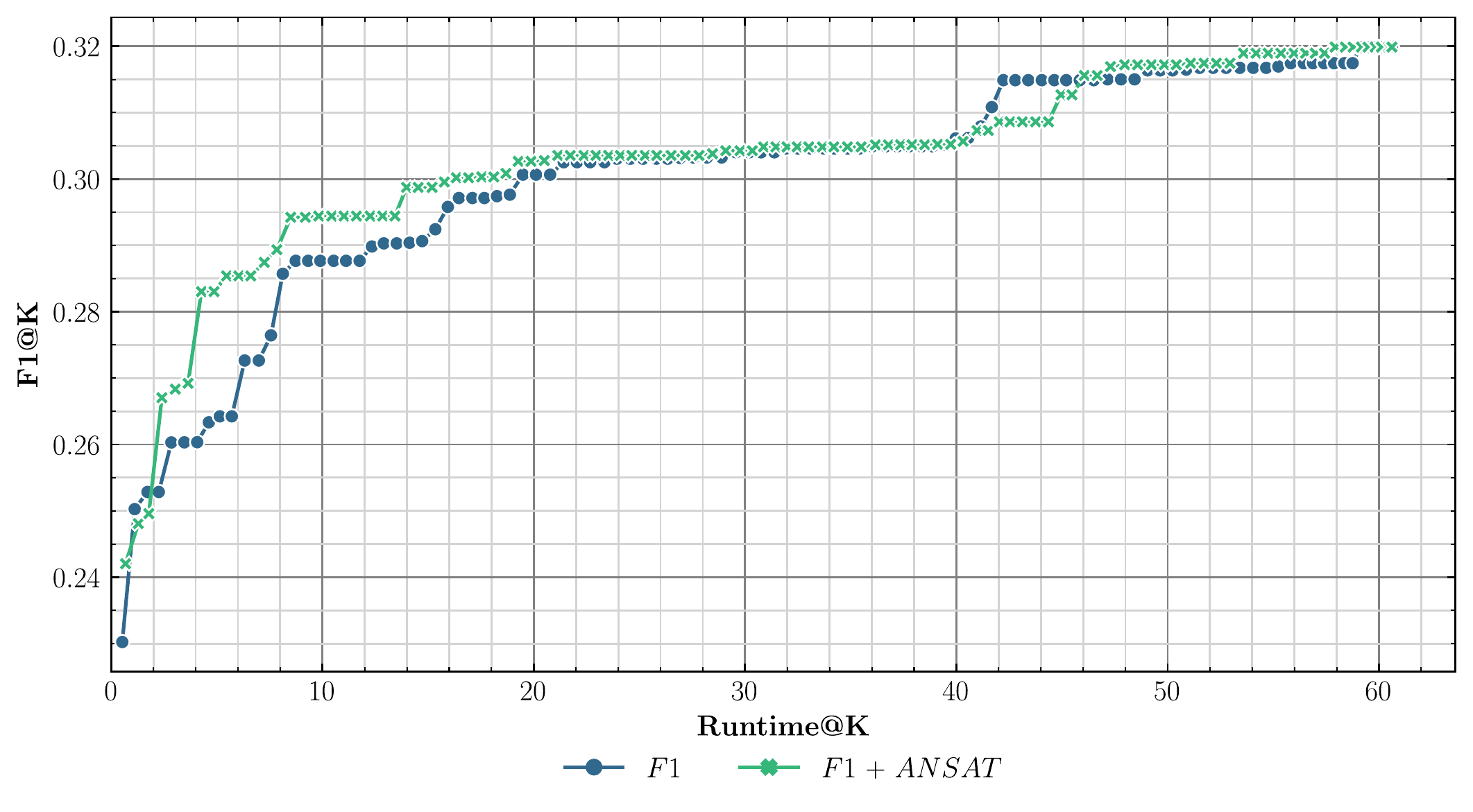}
\vspace{-2mm}
\caption{\small{Performance vs. Runtime results from XGBRegressor. F1 metrics range from 0 to 1, higher is better. Runtime is reported in hours.}}\label{fig:perf_v_runtime}
\vspace{-5mm}
\end{figure}

\begin{definition}
Let $\mathcal{M}$ represent a neural model with layers $l\in L$ and $\mathcal{D}$ represent the collection of positive-class documents (with respect to task sets \small$A$, \small$B$, and \small$AB$) containing words $w\in d \in D$, and with conductance threshold $TAT$ then:

\begin{equation}
    \resizebox{0.91\hsize}{!}{%
    $ANSAT(M_A, M_B, D, TAT)=\sum_{d \in D}\frac{\left(\sum_{w \in d}
    \begin{cases}
        1,  & \text{if } \left(\frac{\sum_{l \in L_{M_A}}conduct(w, l)}{|L_{M_A}|} \ge TAT\right) \text{AND} \left(\frac{\sum_{l \in L_{M_B}}conduct(w, l)}{|L_{M_B}|}  \ge TAT\right)\\
        0,  & \text{otherwise}
    \end{cases}
    \right)}{|D|}$%
    }
\end{equation}\label{eqn:ansat}
\vspace{-5mm}
\end{definition}

\revisions{Through this formulation, we can estimate the pretraining similarity between two tasks (A and B) via their underlying datasets (positive-class documents only) $D_{A}$ and $D_{B}$, as well as the intersection of both, $D_{AB}$. We then use these estimates to predict the effectiveness of a combined model $M_{AB}$ created via transfer learning, i.e. BERT$\rightarrow$Source(A)$\rightarrow$Target(B). In particular, we train an XGBoost~\cite{Chen2016XGBoost} regression model (XGBRegressor) to produce a prediction of the effectiveness of $M_{AB}$, given various feature combinations. We use this model to predict the performance of $M_{AB}$ combinations for each target task B given a set of source tasks A$\in$S, using Positive F1 as our target metric.}


\looseness -1 \revisions{To answer RQ1, we compare the performance predicted by our XGBoost model when using only individual model effectiveness ($M_{A}$ and $M_{B}$ F1-scores) as features vs. those same features + the ANSAT similarity estimations.} \revisions{If active terms as defined by ANSAT are indicative of transfer performance then the XGBoost model with these features should be more effective than the one without}. \ric{Fig.~\ref{fig:rmse_at_k} shows the results of our experiment, reporting} \revisions{RMSE at ranks 10--100 with different feature sets, where $F1$ denotes $M_{A}$ and $M_{B}$ F1-scores and $ANSAT$ denotes the ANSAT scores for $D_{A}$, $D_{B}$ and $D_{AB}$ (under TAT values $[0.05, 0.7]\cap0.05\mathbb{Z}$).}

\revisions{Near the top of the ranking (K=5, 10), we observe the feature set using F1 only to marginally outperform F1 + ANSAT by 1.76\% and 2.76\%, respectively. At ranks K=40 and above, however, we observe that the inclusion of ANSAT results in considerably lower error than using F1 scores alone. Indeed, from these results we can conclude that the overlap of active terms between tasks as measured by ANSAT is valuable evidence when attempting to determine whether the combination of tasks will result in performance gains, answering RQ1.}


\revisions{To answer RQ2, we consider the potential real-world benefits of such performance prediction models when used to reduce task-pair training time. For this experiment, we assume you have a certain budget to train $K$ task-pair combinations and check their performance. For a task, the more combinations you try, the more likely you will find a good combination. As our XGBoost models are predicting which combinations will work well together, we can use this to determine the order of combinations to try, where the goal is to find the best performing combination for each task as early as possible, such that we can end the search early. Fig.~\ref{fig:perf_v_runtime} reports the Positive F1 performance of the best performing model for different depths K, where the x-axis is a conversion of K into the number of hours needed to train that many models for all tasks (Runtime@K).}



\revisions{From the collection of 803 models used as the dataset for our regression model, our best, average performance (F1) was 0.3199, which took 60.6 hours to train. By utilising our regression model, we are able to achieve an F1-score of 0.3003 (only 6.12\% worse than our best-performing F1 model), at only 30 hours or 50.5\% less training time, using the $F1 + ANSAT$ feature space. If we were to accept a 0.034 or 10.78\% reduction in F1-score, we can further reduce our time to 10 hours or a 83.5\% reduction in training time. We note that at lower ranks of K, we observe a consistent increase in performance when including ANSAT in our feature space. At ranks 7, and 10, we observe 8.71\%, and 8.01\% performance increases, respectively, when including ANSAT alongside F1.} Considering these results, there is clearly significant scope for improving performance by leveraging attribution-based techniques, answering RQ2.

\section{Conclusions and Future Work}
In this work, we presented an approach for estimating the suitability for pairs of tasks to be used in transfer learning by comparing their shared, active terms. It is clear that there exists some correlation between term activity and performance, as highlighted by our results. By predicting the projected performance output of each task pair, we managed to achieve up to 83.5\% reduction in training time (for only a 0.034 or 10.78\% reduction in F1). However, while we have demonstrated the value of using conductance to estimate combined model performance pre-training, there is clearly more work needed to increase the accuracy of these estimations, and hence further reduce the space of models that need to be searched. As such, for future work, we propose further analysis into the quantifiable properties that constitute related tasks which could further improve inductive transfer between such tasks.

\bibliographystyle{splncs04}
\bibliography{bibliography}
\end{document}